\documentclass[]{article}
\usepackage[utf8]{inputenc}
\usepackage[english]{babel}
\usepackage[T1]{fontenc}
\usepackage{amsmath}
\usepackage{amsthm}
\usepackage{amsfonts}
\usepackage{amssymb}
\usepackage{listings}
\usepackage{graphicx}
\usepackage{hyperref}
\usepackage{algpseudocode}
\usepackage{framed}
\usepackage{algorithmicx}
\usepackage{algorithm}
\usepackage{subfig}
\usepackage{stmaryrd}
\usepackage{bm}
\usepackage{tikz}
\usepackage{braket}
\usepackage{cite}
\usepackage{wasysym}
\usepackage{wrapfig}
\usepackage[framemethod=TikZ]{mdframed}
\usepackage{xcolor}
\usepackage{pstricks}
\usepackage{faktor}
\usepackage{bbm}
\usepackage{todonotes}
\usepackage[titletoc]{appendix}

\usetikzlibrary{arrows}

\theoremstyle{definition}
\newtheorem{definition}{Definition}[section]
\newtheorem{problem}{Problem}[section]
\theoremstyle{plain}

\theoremstyle{plain}

\newcommand{\vf}{\varphi}
\newcommand{\f}{\mathbb}
\newcommand{\nnsize}[2]{\f R_+^{#1\times #2}}
\newcommand{\size}[2]{\f R^{#1\times #2}}
\newcommand{\wt}{\widetilde}

\newcommand{\di}{\diamond}
\newcommand{\roww}[1]{_{#1,:}}
\newcommand{\coll}[1]{_{:,#1}}

\DeclareMathOperator*{\argmin}{arg\,min}
\DeclareMathOperator*{\argmax}{arg\,max}

\title{Permutation NMF}
\author{Barbarino Giovanni}

\begin{document}

\maketitle

\begin{abstract}
Nonnegative Matrix Factorization(NMF) is a common used technique in machine learning to extract features out of data such as text documents and images thanks to its natural clustering properties. In particular, it is popular in image processing since it can decompose several pictures and recognize common parts if they're located in the same position over the photos. This paper's aim is to present a way to add the translation invariance to the classical NMF, that is, the algorithms presented are able to detect common features, even when they're shifted in different original images.
\end{abstract}

$ $

 Throughout all the document, we indicate the set of nonnegative real numbers as $\f R_+$, and the element-wise (Adamard) product and division between matrices as
 \[
 A \,.*B  \qquad A\, ./B
 \]
Moreover, we'll refer to the $i$-th column and row of a matrix $A$ respectively with $A\coll{i}$ and $A\roww{i}$ .
\section*{Introduction}

The NMF is a powerful tool used in clustering, image processing, text mining, and so on. Its importance grew in the last decade due to its efficacy into extracting meaningful and easily interpretable features out of the data. For example, in the clustering problem of $m$ points into a $n$ dimensional space, the processed data can be naturally viewed as centroids of the clusters, or in its application to text mining, the NMF output clearly points to the common topics touched by the input documents. In this paper the focus is on the applications of NMF to the analysis and decomposition of images, as shown in the article of Lee \& Seung \cite{nature}, where they processed a set of faces and the algorithm automatically recognized their principal features like eyebrows, lips, noses, etc.

A serious drawbacks of this method is that NMF can't recognize the same objects or parts of them if they're located in different places on multiple images, or when they're rotated or deformed. In other words, NMF is not invariant under space transformations, so the input data must always be pre-calibrated and adjusted.

One possible solution may be to add to the dataset a lot of copy of the same image, each time stretched, rotated and shifted in different ways, in order to make the NMF recognize the parts of an image even if they're in different positions and with different shapes, but this leads to an huge rise of input data and of redundancy in the solution. 

Some authors have suggested to set some standard transformations of the images (such as translations or symmetry) and to look for the features we want to obtain, along with additional parameters that indicate for each transformation of each feature if   they're present into the original images. This rises the number of the problem variables by a factor that's usually larger or equal to the number of pixels in a picture, like in \cite{shif} and \cite{shif2}, making the algorithm complexity go up by at least the same factor.

Here is presented a way to attack the problem of the translations, keeping the framework of NMF and the natural the graphical property of its output to represent the wanted parts of images, and bounding the rise in data weight and computational cost  with the number of effective components we want to find and a logarithmic factor.

 In the first chapter we review the original NMF problem, and we'll discuss why it's applicable to image processing. On the second chapter, we introduce the tools and notation needed to state the actual problem we want to solve. On the third chapter, we describe the algorithms used, and derive the asymptotic computational cost. On the fourth chapter we present some experiments on hand-made images, and on the conclusions we'll talk about possible improvements.

 \section{NMF and image processing}
\subsection{Nonnegative Matrix Factorization}
 
 Given a data matrix $A\in \nnsize{n}{m}$ and a natural number $k$, the NMF problem requires to find the matrices $W,H$ that satisfy 
 \begin{equation}\label{NMF}
 \min_{W,H}F (W,H) = \min_{W,H} \|A-WH^T\|^2_F \qquad W\in\nnsize{n}{k}\quad H \in \nnsize{m}{k} 
 \end{equation}
 where we used the Frobenius norm, defined as
 \[
  \|M\|_F^2 = \sum_{i,j}M_{ij}^2
 \]
 
A natural interpretation of NMF derives from the observation that, given any column of $A$, a good solution to the problem finds an approximation of it through a combination of $k$ nonnegative vectors, the columns of $W$, with nonnegative coefficients stored in a row of $H$. This means that the problem is equivalent to find a nonnegative set of $k$ vectors that approximately generate, through nonnegative coefficient, all the columns of $A$ (the minimum parameter $k$ that satisfy such conditions is often referred to as the \textit{nonnegative rank} of the matrix $A$).     
 
Usually, $k$ is much smaller then the other dimensions $n,m$ since the NMF is often used as a low-rank decomposition algorithm, and the resulting columns of $W$, called \textit{features} or \textit{components}, have a meaningful representation as characteristics or parts of the original data, that are the columns of $A$. Moreover, a large value of $k$ implies a large set of exact solutions for the exact NMF problem, and it translates into a lot of local minima into the minimization problem, that leads to inaccuracy on the algorithmic part, and ambiguity in the interpretation of solutions.

 An other feature that is usually required to the input data is the sparsity, since it is proved that can improve the quality and understandability of the solution, along with gaining uniqueness properties (for further studies, see \cite{sparse}, that proposes a preprocessing to improve the sparseness of $A$). 
 
 A common way to take advantage form the non-uniqueness of the solution is to normalize rows and columns of $A,W,H$, through a positive diagonal matrix $S$ of dimensions $k\times k$. In fact, given any pair $(W,H)$, then $WH^T=WS^{-1}SH^T$, so the matrices $(W',H') = (WS^{-1},HS^T)$ are still nonnegative, and this transformation doesn't change the error we want to minimize. If we set the diagonal of $S$ as the $l^1$ norm of the columns of $W$, then $W'$ is column stochastic, and if the input matrix $A$ is also column stochastic, then an exact solution $A=WH^T$ requires the columns of $H^T$ to be stochastic as well, so that the columns of $A$ belong to the convex hull generated by the features in $H$.
 
 We now see how this considerations are important in practical applications.
 
 \subsection{Image Processing}
 
 One of the problem confronted by researchers in image processing is to decompose different images into common parts or features, both for identification purposes or for compression ones. For example, a common technique used in animation in order to contain the memory used is to not memorize into digital supports every pixel of each single frame, but to memorize only particular compressed or coded informations that lets a recorder to reproduce the film with little loss of quality. 
 
 In general, when confronted with a large set of images like the frames of a film, or a database of similar pictures, it can be convenient to memorize the common parts only one time, gaining space and also computational time for the recombining process. The problem is thus to find an efficient algorithm that automatically recognizes the common features and an intelligent way of storage of the informations.
  
  
 Given a gray-scale image $M$ expressed as a matrix of pixels, with values in the real range $[0,1]$, we can transform it into a real vector with as many coordinates as the pixels in the image. In particular, if $M\in\nnsize{r}{s}$, then we stack the columns of the matrix on top of one another, and obtain the vector $v\in \f R_+^{rs}$ defined as
 \[
  v_{i + (j-1)r} = M_{ij} \quad \forall\, i,j
 \] 
 Given a set of pictures $\set{M_i}_{i=1:m}$ of the same shape, we can now vectorize them and stack the corresponding vectors as the columns of our data matrix $A$, and if we call $n=rs$ the number of pixels of a single picture, $A$ becomes a nonnegative matrix in $\nnsize{n}{m}$, so, after having fixed the number $k$ of common component we want to find, the NMF framework produces two matrices $W,H$ such that $A\sim WH^T$. 
 
 As already noticed, each column of $A$ is approximated by a linear combination of the columns of $W$, that are nonnegative vectors of length $n=rs$. After having normalized $W$ by multiplication with a diagonal positive matrix (as discussed above), we can see its columns as images in the shape $r\times s$, so a generic column of $A$, that is one of the original images, is now approximated as the superimposition of the pictures represented by some of the columns of $W$. 
 \[
 A\sim WH^T\implies A_{:,i} \sim H_{i,1}W_{:,1} + H_{i,2}W_{:,2} + \dots + H_{i,k}W_{:,k} 
 \]     
Ideally, the images in $W$ are parts of the pictures in $A$, like localized objects in the 2D space, so they're usually sparse and disjoint images, that translates into sparse and nearly orthogonal vectors. In a famous experiment, Lee \& Seung \cite{nature} processed a set of faces and the NMF automatically recognized their principal features like eyebrows, lips, noses, eyes, and so on, so that they were immediately human-recognizable. This example shows the importance of NMF as a decomposition tool for graphical entities. 

As already said, the sparseness and the choice of $k$ are important factors. The sparseness is an index of the uniqueness of the solution, that is important on the side of interpretation of the output, since different solutions usually brings up set of pictures not human-recognizable as real objects and features. On the side of compression, we can see that the original $nm$ pixels of $A$ are now coded into $kn$ pixels in $W$ and $km$ coefficients in $H$, so the compression is useful when the approximation is good with a low $k$. On terms of images, it means that there are few components that span the whole set of pictures.   
 
\subsection{Transformations Issues}

When we use NMF on a matrix $A$ we usually expect the original images to have some predominant common features, so that the algorithm can find them with little noise. This may be true in the case of sets of static pictures, when calibrated and centered, but even in the case of facial recognition, there may be cases of misalignment, as already noticed by \cite{gaussinv} and many others. In general, the NMF suffers in this cases since it is not invariant under a vast set of transformations, for example shifts, rotations, symmetry, stretches and so on, in fact the common features must be in the exact same positions on the different pictures in order to be pinpointed. 

This is a common problem faced in the animations programs, since, even if the subjects in a scene of a footage are the same, they constantly move on the screen, so their detection must follow some temporal scheme, and can't be performed by a simple NMF.

Possible ways to deal with this problem are to change the data in one of the three matrices $A,W$ or $H$. For example, if we add ta $A$ a transformed copy of each original picture for every transformation in a set we choose, then the common features get detected even if they're deformed, but this increases the size of the problem by the square of the number of alterations used, that's usually greater than the number of pixels in a single image. One possible solution is obviously to rise the parameter $k$, but this leads to instability in the solution, as we already discussed.

A good idea seems instead to rise the quantity of data contained in the matrix $H$, since we strife to maintain the graphical property of the columns of $W$ to represent the common features of the original images. In the next chapters we'll define new notations and operators to deal with a matrix whose elements are capable to transmit more informations on pictures than simple real numbers.

 \section{Permutations}
 
 In this document, our focus is on the problems related to the lack of translation invariance of NMF, so we'll use shift permutations to modify the kind of elements contained in the matrix $H$. First of all, we define an operator between matrices not necessarily real.

  \subsection{Diamond Operator}
  
  Given an element $\tau \in \f R\times S_n$, represented by a couple $\tau \equiv [r,\sigma]$, where $r$ is a real number, and $\sigma$ is a permutation of $n$ indexes (that is, an element of the permutations group $S_n$), then it's well defined its action on a real vector $v\in\f R^n$
  \[
  \tau(v) \in \f R^n \,\,:\,\, \tau(v)_i = [r,\sigma](v)_i = rv_{\sigma(i)} \quad\forall\,i
  \]
  The action of $\tau$ on $\f R^n$ makes it a linear operator, so it can be represented by a matrix, and in particular, since the action of each permutation $\sigma\in S_n$ is associated with a permutation matrix $P_\sigma$, it's easy to see that 
  \[
  \tau \equiv  [r,\sigma] \implies \tau(v) = rP_\sigma(v) 
  \]

  The algebra generated by the permutation group over the real field is denoted as $\f RS_n$, and its elements are finite sums of $\f R\times S_n$ elements
  \[
  \alpha \in \f RS_n \implies \alpha= \sum_{i=1}^s [r_i,\sigma_i] \quad r_i\in\f R\quad \sigma_i\in S_n\,\,\forall i
  \]
  As before, these elements have a natural action on $\f R^n$, that is an extension of the action of $\f R\times S_n$, given by
  \[
  \alpha(v) = \sum_{i=1}^s [r_i,\sigma_i](v) = \sum_{i=1}^s r_iP_{\sigma_i}v = \left(\sum_{i=1}^s r_iP_{\sigma_i}\right) v
  \]
  so there exists an homomorphism of $\f R$ algebras $\vf :\f RS_n\to \f R^{n\times n}$ that associates to each element of the algebra a real matrix, and later we'll see how it behaves on a particular subgroup. \\

  Let's now suppose that $N$ is a matrix with entries in the above described algebra $\f RS_n$, and $M$ is a real matrix. We need an operator to apply the elements of $N$ to the columns of $M$ , so we define the \textit{diamond product} :
  \begin{definition}[Diamond Product]
  The diamond operator between a real matrix $A\in\size{n}{m}$ and a matrix $N\in (\f RS_n)^{m\times k}$ is defined as
  \[
  (A\di N)\coll{i} := \sum_j N_{ji}(A\coll{j})
  \]
  and returns a real matrix in $\size{n}{k}$.
  \end{definition} 
  In other words, the $i$-th column of the diamond product is a linear combination of permutations of $M$ columns, with coefficients and permutations described by the elements of the $N$'s $i$-th column.
  
  Let's also define the multiplication between two matrices with entries in the algebra of permutations. Remember that $\f RS_n$ is an algebra, so sum and product are well defined, and the elements of $\f RS_n$ can be viewed as well as matrices through the homomorphism $\vf$, so the two operations correspond to the usual sum and composition of matrices. 
  \begin{definition}[Diamond Product]
    The diamond operator between two matrices $M\in (\f RS_n)^{n\times m}$ and $N\in (\f RS_n)^{m\times k}$ is defined as
    \[
    (M\di N)_ {ij} := \sum_k N_{kj}\cdot M_{ik}
    \]
    and returns a matrix in $(\f RS_n)^{n\times k}$.
   \end{definition} 
  This operation differs from the normal multiplication of matrices only because $\f RS_n$ isn't a commutative algebra, so we need to specify the order of the multiplication between the elements. The inverted order is necessary to partially maintain the associativity of the operation: given a real matrix $A$, and two matrices $N,M$ with elements in the algebra, it's easy to verify that
  \[
  (A\di M)\di N = A\di (M\di N)
  \]
  Ideally we need to invert the elements of $N$ and $M$ since $M$ is the first to act on the columns of $A$, followed by $N$.
  
  One downside of this operation is that it doesn't cope well with the normal matrix multiplication: given $A,B$ real matrices, and $M$ a matrix in the permutation algebra, then
  \[
  A(B\di M) \ne (AB)\di M
  \]
  
  Let's now return to image transformations, and focus on a particular subgroup of the permutation algebra.
  
 \subsection{Shifts and Circulant Matrices}
 
 Given a gray-scale image $M$, we've seen how to transform it into a  vector $v\in \f R_+^{rs}$. We want now to codify a shift on the image as a vectorial transformation:
 a shift of the original image $A$ by $r_1$ position on the horizontal axis and $s_1$ position on the vertical one will be encoded as a circular shift on $v$ of magnitude $p = r_1r + s_1$, that is, we produce a vector $w$ whose $i$-th coordinate is the $(i+p)$-th coordinate of $v$. 
 
 If we call $n=rs$, we can denote as $T_n$ the cyclic subgroup of the permutation group $S_n$ whose elements shift cyclically all the indexes of vectors in $\f R^n$ by an integer constant. We'll call $\sigma_p$ the shift by $p$ position, where $p\in \faktor {\f Z}{n\f Z}$:
 \[
 \sigma_p \in T_n \quad v\in \f R^n\quad p\in \faktor {\f Z}{n\f Z}  \implies \sigma_p(v) = w \quad : \quad w_i = v_{i+p} \quad \forall i
 \] 
 where the indexes are to be considered modulus $n$.

 The elements of $T_n$ are linear operators, so can be represented by $n\times n$ matrices through the above mentioned homomorphism $\vf$. In particular, the element $\sigma_1$ is associated to the circulant matrix $C$ that has 1 on the first cyclic superdiagonal and 0 anywhere else, and $\sigma_p = \sigma_1\circ \dots\circ \sigma_1$, so $\vf(\sigma_p) = \vf (\sigma_1)^p = C^p$ that has 1 on the $p$-th cyclic diagonal and zero otherwise. 
 \[
 \vf(\sigma_1) = C = 
 \begin{pmatrix}
 0 & 1 &   &  &   \\
   & 0 & 1 & \phantom{\ddots}  &    \\
   &   & 0 & \ddots &   \\ 
   &   &   &     \ddots &  1\\
 1 &   &   &   \phantom{\ddots}  & 0
 \end{pmatrix}
 \qquad 
 \vf(\sigma_2) = C^2 = 
 \begin{pmatrix}
 0 & 0 & 1 &    &   \\
   & 0 & 0  & \ddots &   \\ 
   &   & 0  & \ddots &  1\\
 1 &   &    & \ddots &  0\\
 0 & 1 &    & \phantom{\ddots}  & 0
 \end{pmatrix}
 \quad \dots
 \]
 \[
 \sigma_p(v) = C^p v
 \]
 In the next section, we'll use elements of type $\tau = [r,\sigma_p]\in \f R_+ \times T_n$ to define a new problem with the same shape of a normal NMF, but on different domains, and since the shift $\sigma_p$ is completely identified by the remainder class $p$, we'll refer to $\tau$ as the couple $[r,p]$.
 %
 %
 

 %

 \subsection{PermNMF}

 Now we reconsider the classic NMF, and widen the domain of the matrix $H$. Our aim here is to find a new method to decompose pictures into common components, even when they're shifted, so, like in the NMF, we stack the original images as columns of the matrix $A$, and look for a matrix $W$ whose columns are the wanted common features, and a matrix $H$ with elements in $\f R_+\times T_n$, so that it can tell us both the intensity and the position of each component in $W$ into each original picture in $A$. 
 
 In particular, we want to rewrite the NMF problem as

 \begin{problem}[PermNMF]
 Given a matrix $A$ is in $\nnsize{n}{m}$, we want to find a matrix $H$ in $(\f R_+\times T_n)^{m\times k}$ and a matrix $W$ in  $\nnsize{n}{k}$ that minimize
 \[
 F(W,H) = \| A - W \di H^T\|_F^2   
 \]
 \end{problem}
 
 The diamond operator is defined on elements of $\f RS_n$, but we restrict the entries of $H$ to elements in $\f R_+\times T_n$, so that a single image (column of $A$) is a linear combination of the images represented by the columns of $W$, but shifted. We notice that expanding further the domain of $H$ usually leads to trivial and useless solutions; for example, if we let the elements of $H$ be in $\f R_+T_n$, that are linear nonnegative combinations of permutations in $T_n$, then even with $k=1$ there's a trivial solution that decomposes perfectly the matrix $A$: 
 \[
 A = W\di H^T \qquad W = e_1 \qquad H_{i,1} = \sum_j [A_{ij}, j-1] 
 \] 
 in fact,
 \[
 (W\di H^T)\coll{i} = H_{i,1}(W) =  \sum_j A_{ij} \sigma_{j-1}(e_1) =  \sum_j A_{ij} e_j = A\coll{i} 
 \]
 In other words, a linear combination of the translations of a single pixel can reconstruct any image, so it is an exact and completely useless solution.
 Moreover, expanding to the group $T_n$ usually leads to the dismembering of the images represented by the columns of $W$, so we stick to work with this framework for this document.
 
 An other particularity of this formulation is that, if we impose that each element of $H$ must be of the type $[r,0]$, that is, we fix all the permutations to be the trivial identity, then the problem returns exactly the original NMF, and the diamond operator coincides with the normal matrix multiplication.
 
 \section{Algorithm}
 The PermNMF has the same structure of the normal NMF, so we can try to use similar solving algorithms. A characteristic we'd want from our solution is the sparsity of the $W$ columns, since they should represent isolated objects in the images, so the first algorithm considered is the MU update, since it is known to naturally produce sparse solutions. Unfortunately, the MU method efficiency, in the NMF case, comes from the approximation
 \[
 W^T A \sim W^T(WH^T) = (W^TW) H
 \]  
 but in our case, as already stated, there's no associative property
 \[
 W^T A \sim W^T(W\di H^T) \ne (W^TW)\di H^T
 \]
 For this reason, we resort to an ALS/PG setting.
  \begin{framed} 
  \textbf{ALS Adapted Update Method} 
  
  \textit{Inputs : $  A\in \nnsize{n}{m}, \quad  W\in \nnsize{n}{k}, \quad H\in (\f R_+\times T_n )^{m\times k} $ }
     \begin{algorithmic}
      \State $H =\argmin_{X\in (\f R_+\times T_n)^{m\times k}} \|A-W\di X^T\|^2_F$
      \State $W =\argmin_{X\in \size{n}{k}} \|A-X\di H^T\|^2_F$  
      \State make $W$ nonnegative
     \end{algorithmic}
  \end{framed}
 
 The update of $W$ requires to solve a convex problem, so we can use some of the usual methods, like a modified Projected Gradient; this one is particularly good for this case, since we can't transpose the expression in order to obtain the setting of the Active Sets algorithms. 
 
 For simplicity, we use the following PG algorithm, where we stop in case of low error or small step:
  \begin{framed} 
  \textbf{PG Update Method}
   
  \textit{Inputs : $  A\in \size{n}{m}, \quad  W\in \size{n}{k}, \quad H \in(\f RS_n)^{m\times k},\quad iter\in \f N$  }
     \begin{algorithmic}
 	 \For {$i=1:iter$}
 	     \State $W = W - \nabla_W F(W,H)/i$
 	     \State $err = \|A-W\di H'\|$
 	     \If {$err<.001$ or $\|\frac{\partial F(W,H)}{\partial W}\|<.001$}
 	     	     	\State \textbf{break}
 	     \EndIf
 	 \EndFor
 	 \State \Return $W$
     \end{algorithmic}
  \end{framed}
 We'll refer to this function from now on as
 \[W = PG(A,W,H)\]
 The computation for the gradient in the algorithm are developed in Appendix A, and it shows that
 \[
 \nabla_W \|A-W\di H^T\|^2_F = -2(A-W\di H^T)\di H'
 \]
 The operations performed in each cycle of the method have a computational cost of $O(mnk)$.
 
 Let's now focus on the update of $H$, that requires to solve an optimization problem on the group $\f R\times T_n$. We start by solving a largely simplified problem.
 
 \subsection{Single Permutation NNLS}
 
 Let's suppose to have two vectors $v,w$ in $\f R^n$, and we want to find the best element $\tau = [r,p]$ of $\f R_+\times T_n$ that minimizes
 \[
 E(\tau) = \|v - \tau(w)\|^2
 \]
 where the norm used is the euclidean one.
 
 A natural assumption is that $w\ne 0$, otherwise every element $\tau$ gives the same value of $E(\tau) = \|v\|^2$.
 If we knew the optimal $p$, then we could find $r$ without fail, because it becomes a simple Nonnegative Least Squares (NNLS) problem.
 \[
 r_p := \argmin_{r\in \f R}\|v - r\sigma_p(w)\|^2 =  
 \faktor{v^T\sigma_p(w)}{\sigma_p(w)^T\sigma_p(w)}
 \]
 \[
 r_p^+ := \argmin_{r\in \f R^+}\|v - r\sigma_p(w)\|^2 =  
 \begin{cases}
 0 & v^T\sigma_p(w)< 0 \\
 \faktor{v^T\sigma_p(w)}{\sigma_p(w)^T\sigma_p(w)} & v^T\sigma_p(w) \ge 0
 \end{cases}
 \]
 A simple solution consists into computing the optimal $r_p^+$ for every $\sigma_p\in T_n$, and check which couple $[r_p^+,p]$ gives us the minimal error. We know that $\sigma_p(w)^T\sigma_p(w)=\|w\|^2$, so we can compute the error as a function of $p$
 \[
 \|v - r_p\sigma_p(w)\|^2 = \|v\|^2 - \frac{(v^T\sigma_p(w))^2}{\|w\|^2}
 \]
 The problem is thus equivalent to maximize $(v^T\sigma_p(w))^2$, but we're interested only in the positive case, so we focus on maximizing the scalar product $v^T\sigma_p(w)$, since if $v^T\sigma_p(w)<0$ then $r_p^+ = 0$ for every $p$, so $E([r_p^+,p]) = E([0,p])=\|v\|^2$.\\
 
  By definition, $\sigma_p(w)$ is the vector $w$ shifted, so we can call $C$ the real nonnegative matrix that has all the shifted versions of $w$ as columns, and compute the maximal component of $v^TC$. Since $C$ is a circulant matrix, this operation costs $O(n\log n)$ if performed with Fast Fourier Transformations, so this method is fast and gives us the correct solution.

 \begin{framed} 
  \textbf{Single Permutation NNLS}
   
  \textit{Inputs : $  v,w\in \f R^{n},\quad w\ne 0$  }
  
  \textit{Output : $  \tau\in \f R^+\times T_n$  }   
     \begin{algorithmic}
 	 \State $p = \argmax_i \,\,(v^TC)_i$
 	 \If {$v^T\sigma_p(w)>0$}
 	 	\State $r = v^T\sigma_p(w) / \|w\|^2$
 	 	\Else 
 	 	\State $r=0$
 	 	\EndIf
 	 \State \Return $[r,p]$
     \end{algorithmic}
  \end{framed}
 
\noindent From now on, we'll use this algorithm with the syntax $$\tau = \text{SinglePermNNLS}(v,w)$$
Let's now increment the number of permutations needed.
 
 \subsection{Multiple Permutation NNLS}

 Given now a vector $v\in \f R^n$, and a bunch of vectors $w_1,w_2,\dots,w_k\in \f R^n$ we can now try to find the best elements $\tau_1,\dots,\tau_k\in(\f R_+\times T_n)$ that minimize the quantity
 \[
 \|v - (\tau_1(w_1) + \tau_2(w_2) + \dots + \tau_k(w_k))\|
 \]
 We're thus looking for the best linear combination with positive coefficients of the shifted vectors $w_i$ that gives us the original vector $v$.
 If we call $W$ the matrix with $w_i$ as columns, and $x$ the (column) vector of $\tau_i$, then we can rewrite the problem in a compact way as 
 \[
 \min_{x\in (\f R\times T_n)^k} \|v-W\di x\|^2  \qquad v\in\f R_+^n \quad
 W \in \f R_+^{n\times k}
 \]
 A way to solve this problem is using the precedent algorithm in an alternated fashion. In fact, if we fix $\tau_2,\tau_3,\dots,\tau_k$, then it becomes a Singular Permutation NNLS problem on $\tau_1$, and we know how to solve it exactly.
 
 So we can solve the problem sequentially for each $\tau_i$ and repeat. The initial value of $x$ is usually given as an input parameter, but it can also be generated casually at the beginning of the algorithm.
 
 \begin{framed} 
  \textbf{Multiple Permutations NNLS}
   
  \textit{Inputs : $  v\in \f R^{n},\quad W\in \f R^{n\times k},\quad iter\in \f N, \quad x\in (\f R^+\times T_n)^k$  }
  
  \textit{Output : $  x\in (\f R^+\times T_n)^k$  }   
     \begin{algorithmic}
     \State $w = W\di x$
 	 \For {$j=1:iter$}
 	 \For {$i=1:k$}
 	 	\State $w = w - x_i(W\coll{i})$
 	 	\State $x_i = SimplePermNNLS(v-w,W\coll{i});$
 	 	\State $w = w + x_i(W\coll{i})$
 	 \EndFor
 	 \EndFor
 	 \State \Return $x$
     \end{algorithmic}
  \end{framed}
  From now on, we'll use this algorithm with the syntax $$x = \text{MultPermNNLS}(v,W,x)$$
 Its computational cost is the number of iterations multiplied $k$ times the cost of The Single Permutation Problem, so it is $O(kn\log(n)) $ considering $iter$ as a constant. In particular cases, it may be useful to randomize the choice of the index $i$, since it's important not to impose a preference order on the components in $W$. 
 
 \subsection{Final Method}
 We can now return to the original problem
 \[H =\argmin_{X\in (\f R_+\times T_n)^{m\times k}} \|A-W\di X^T\|^2_F\]
 Like the normal NMF, it can be decomposed into smaller problems
 \[\|A-W\di X^T\|^2_F=\sum_{i=1}^m  \|A\coll{i}-W\di (X^T)\coll{i}\|^2\]
 \[
 H\roww{i} = \argmin_{x\in (\f R_+\times T_n)^{k}} \|A\coll{i}-W\di x\|^2
 \]
 that can be solved with the Multiple Permutation NNLS algorithm. If we put everything together, we obtain the final method
 
  \begin{framed} 
  \textbf{ALS Adapted Update Method} 
  
  \textit{Inputs : $  A\in \nnsize{n}{m}, \quad  W\in \nnsize{n}{k}, \quad H\in (\f R_+\times T_n )^{m\times k} $ }
     \begin{algorithmic}
     \For {$i=1:m$}
      \State $H\roww{i} =MultPermNNLS(A\coll{i}, W, (H^T)\coll{i})$
      \EndFor
      \State $W = GD(A,W,H)$  
      \State make $W$ nonnegative
     \end{algorithmic}
  \end{framed}
 
 Every step of This ALS Update Method costs $O(kmn\log(n))$ if we consider the number of iterations in the internal methods as constants. We will stop the updates when the convergence is too slow, when we loop on the same matrices, or when we reach a number of iteration too high.
 
 \subsection{Extension and Other Works}
 
 Given a set a pictures, now we're able to perform a PermNMF and obtain a set of $k$ common features that can reconstruct the original data once combined through coefficients and permutations codified in $H$. Given one of the images in $W$, the algorithm tells us if it is present in the original images, but it doesn't detect if it appears multiple times.
 One example of  such instance may be a set of radar images, in which different objects intercepted by the wave signals have distinct shapes, but each one can appear multiple time in the same picture.
 
 One possible solution is to perform an initial PermNMF with a parameter $k$ proportional to the effective number of distinct objects with multiplicity that can appear on a single image, discard the found components with low coefficients, and repeat the PermNMF on the output components with a low $k$ corresponding to the number of distinct shapes without multiplicity. Let's call $K$ the first larger parameter, and $A\in \nnsize{n}{m}$ the set of pictures to analyze. We obtain
 \[
 A \sim \wt{W}\di H_1^T \sim (W \di H_2^T)\di H_1^T = W \di (H_2^T \di H_1^T)
 \]
 where $\wt W\in \nnsize{n}{K}$, $W\in \nnsize{n}{k}$ and $H_1\in (\f R_+\times T_n )^{m\times K}$, $H_2\in (\f R_+\times T_n )^{K\times k}$, so the final decomposition will be again a real matrix with $k$ components, and a matrix $H_2^T \di H_1^T\in (\f R_+T_n )^{k\times m}$. This last matrix is able to tell, for each component, even if there are multiple instances in every original image.
 
 The computational cost of such method (for each cycle, till convergence) is \[O(nmKlog(n) + nKklog(n)) = O(nKlog(n)(m+k))\] that, under the assumption $k<<m$, is equivalent to $O(nmKlog(n))$, meaning that the second step has a negligible computational cost compared to the first. If $K$ is still on the order of magnitude of $k$, the asymptotic cost doesn't change, but if that's not the case, it is better to look for other ways.\\
 
 On this topic, Potluru, Plis and Calhourn in \cite{shif} offer an algorithm that uses Fast Fourier Transformations and circulant matrices in order to compute and codify permutations of the components, called ssiNMF (sparse-shift invariant NMF). As in the PermNMF, the basic idea is to find $k$ components and a set of permutations that could reconstruct the original images, but the ssiNMF sets as target the permutations in the group $\f R_+T_n$, corresponding through $\vf$ with all the circulant nonnegative matrices, so that all the operations can be performed through FFTs. Thanks to this, their algorithm is able to directly construct an approximation
 \[
 A \sim  W\di H^T  \qquad W\in \nnsize{n}{k} \quad H\in (\f R_+T_n )^{m\times k}
 \]
 Eggert, Wersing and Korner in \cite{shif2} took a more general approach to the problem: as we set a subgroup of $S_n$, they chose a general set of transformations of the plane, seen as operators on the columns of $W$, and multiplied the number of parameter of $H$ by the cardinality of the chosen set, so that for each transformation of the components there would be coefficients in $H$ stating their intensity in the original images.
 
 Both the approaches suffer by the presence of the trivial and exact solution described in section 2.3: a single pixel can generate any image if we allow too many transformations of the space. They propose to perform a common modification on the NMF framework, that is adding a penalty factor to ensure the sparseness of the output, since the presence of a single pixel in the component output corresponds to a lot of positive coefficients in $H$, and it leads to the presence of an additional parameter $\lambda$ to set manually or through validations techniques. 
 
 An other characteristic of both the algorithm is the rise in memory used and asymptotic computational cost by at least a factor on par with the number of pixels on a single image, leading to a cost by iteration at least of $O(n^2mk)$. When compared with the PermNMF algorithm, we see that they're comparable when $K\sim nk/\log(n)$, meaning that a component have to appear in the original image on average $n/\log(n)$ times.

 \section{Experiments}

 In these experiments, we use the PermNMF algorithm seen in the previous chapter, with the initial parameters $W$ and $H$ generated randomly, and the $iter$ variable set to 10 in both the $MultPermNNLS$ and the $PG$ methods.\\
 
 In the first experiment (Figure 1) we use 2 simple shapes (a square and a cross) of 9 pixel that move into a frame of dimensions 20x20, and add a casual error of mean 0.15 (where each pixel has an intensity between 0 and 1). In this case the algorithm manages to find the right components after less than 10 repetitions on average. The images shown on the bottom row are the column of $W$, and they're distinguishable as a cross and a square, with little noise given by the imperfections on the original images. \\
 
 In the second experiment, we generate 20 images of shape 30x30 from three simple figures (a plane, a tank and a ship), with a nois of mean 0.15. Each image can include up to two copies of the same figure, so we need to perform a first PermNMF with $k=6$, and then a second time with $k=3$ to extract the original ones. The first application of the algorithm is slowed down by the presence of the same shapes multiple times in the images, but the second application is real fast. As said, we managed to extract first the common features with their multiplicity, and then the actual features. Multiplying the two $H$ matrices we obtained in the two steps of the algorithm, we can deduce the actual position with multiplicity of the shape found in all the 20 original images.

 \begin{figure}[h]
  \includegraphics[scale=.3]{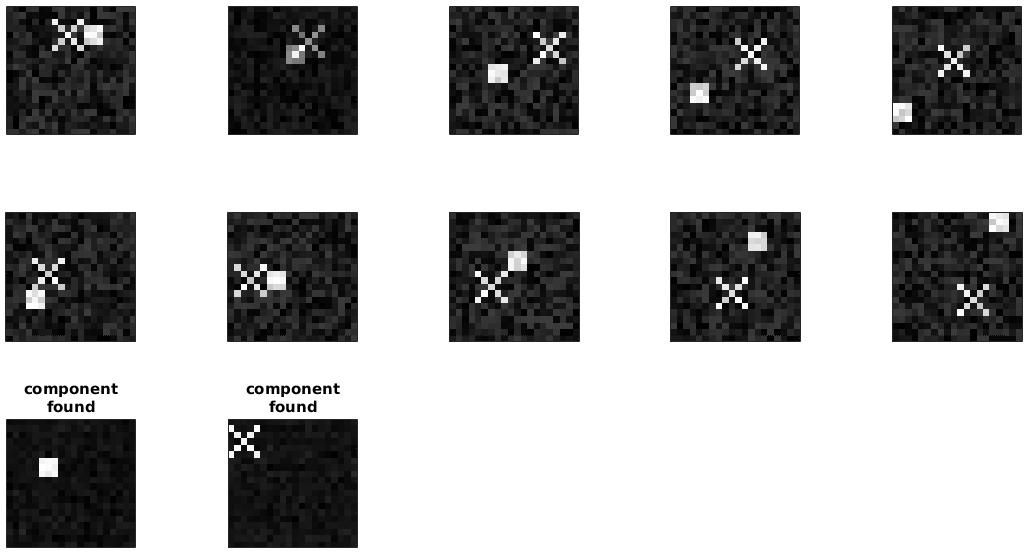}
  \caption{On the first 2 rows, there are the original 10 images, that are the columns of $A$. The other 2 rows are the components found as columns of $W$.}
  \end{figure}

  \begin{figure}[h!]
   \includegraphics[scale=.35]{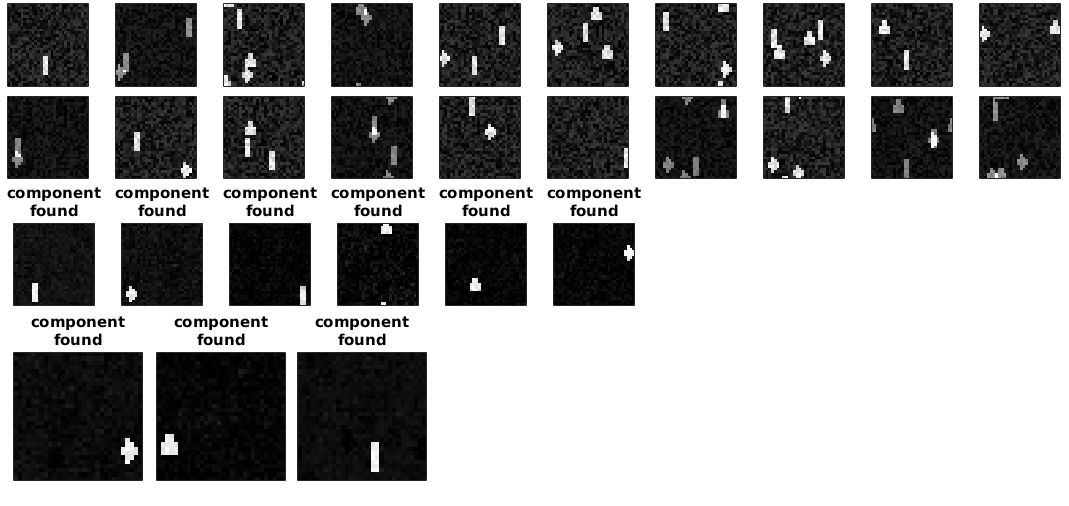}
   \caption{On the first 2 rows, there are the original 20 images, composed by three base pictures translated and superimposed. on the third row there are the components found by the first PermNMF, and in the last row there is the final output of the second PermNMF, that coincide with the base pictures.}
   \end{figure}
     
 \newpage 
 \section{Future Works}
 
 The PermNMF has not been throughly studied and analyzed. First of all, it lacks a convergence result, both because the usual arguments used for the ALS algorithms vastly use the fact that the two subproblems in the classical NMF are convex, and because we switched the framework to non-continuous spaces such as $\f R_+ \times \faktor{\f Z}{n\f Z}$, where it is still not even well defined a canonical concept of "local minimum" (the usual topological embedding of this space in $\f R^3$ gives a notion of stationary points that doesn't cope well with the nature of permutations). 
 
 On the point of view of the PermNMF problem, there's a lot to say, for example, on whether there exists an exact algorithm, or if there are bounds on the minimum $k$, or even if the solution is unique (up to trivial transformations). In \cite{sparse}, Gillis find a preprocessing for the input data $A$ that gives a more well-posed problem then the normal NMF, so such a transformation could be beneficial even to the PermNMF. In \cite{NPhard}, the authors found precise conditions for $A$ under which there exists a polynomial time algorithm for the exact NMF problem, and stated that in general the approximation problem is NP-hard, so it's highly possible that even the PermNMF problem is a NP-hard problem, and that a the polyomial time algorithm could be adapted for this case.
 
 On the side of the algorithm itself, it's possible that a MU (Multiplicative Update) approach on $W$, even if expensive, could retain its descend property, so it can become a substitute or an aid for the PG method. On both the update of $W$ and $H$, it is still possible to apply a CD (Coordinate Descend) method, even if it also lost most of his efficiency due to the bad behavior of the diamond operator. Both this methods, MU and CD, are also recommended for the generation of sparse solutions, a feature we'd like to obtain. On the Multiple PermNNLS algorithm, moreover, it's also possible to consider an active-set like method to choose preemptively which element to update in every cycle, in order to make the error drop faster.
 
 Eventually, we studied the problem when the elements of $H$ are restricted to $\f R_+\times T_n$, but it's possible also to consider other subgroups and subalgebras of $\f RS_n$ in order to encode different transformations of the plan, or just to make the NMF invariant with respect to particular linear operators.

 \begin{appendices}
   \section{Computation of $W$ gradient}

 Let's compute the gradients needed.
 \[
  \|A-W\di H^T\|^2_F = \sum_{i,j} \left[a_{ij} - \left(\sum_s h_{js}(w_s)\right)_i \right]^2
 \]
 In the following steps, we consider the general element of $H$ as a (circulant) matrix, using implicitly the homomorphism $\vf$.
 \[
 \frac{\partial}{\partial w_{uv}} \sum_{i,j} \left[a_{ij} - \left(\sum_s h_{js}(w_s)\right)_i \right]^2
 \]
 \[
 = 
 -2 \sum_{i,j} (A-W\di H^T)_{ij}\frac{\partial}{\partial w_{uv}} \left(\sum_s h_{js}(w_s)\right)_i 
 \]
 \[
 =
 -2 \sum_{i,j}(A-W\di H^T)_{ij}   \frac{\partial}{\partial w_{uv}}(h_{jv}(w_v))_i
 \]
 \[
 =
 -2 \sum_{i,j}(A-W\di H^T)_{ij}   \frac{\partial}{\partial w_{uv}}\sum_k (h_{jv})_{ik}w_{kv}
 \]
 \[
 =
 -2 \sum_{i,j}(A-W\di H^T)_{ij}   \frac{\partial}{\partial w_{uv}} (h_{jv})_{iu}w_{uv}
 \]
 \[
 =
 -2 \sum_{i,j}(h_{jv})_{iu}(A-W\di H^T)_{ij}   
 \]
 If we denote the matrix $h_{jv}$ as the couple $[r,\sigma_t]$, then its transpose is represented by the couple $[r,\sigma_{n-t}]$. Let's call $H'$ the matrix with the same dimension of $H$ and $h_{ij}=[r,\sigma_t] \implies h'_{ij}=[r,\sigma_{n-t}]$, so we have
 \[
 (h_{ij})_{hk} = (h'_{ij})_{kh}
 \]
 We can continue the computation as
 \[
 -2 \sum_{i,j}(h_{jv})_{iu}(A-W\di H^T)_{ij}  
 \]
 \[
 =
 -2 \sum_{i,j}(h'_{jv})_{ui}(A-W\di H^T)_{ij}
 \]
 \[
 =
 -2 \sum_{j}(h'_{jv}(A-W\di H^T)_{j})_u
 \]
 \[
 = -2((A-W\di H^T)\di H')_{uv}
 \]
 So we can write in a compact form the gradient
 \[
 \nabla_W \|A-W\di H^T\|^2_F = -2(A-W\di H^T)\di H'
 \]

 \end{appendices}

\bibliography{mybib}

\bibliographystyle{plain}

\end{document}